\documentclass[letterpaper]{article} 
\usepackage{aaai24} 
\usepackage{times}  
\usepackage{helvet}  
\usepackage{courier}  
\usepackage[hyphens]{url}  
\usepackage{graphicx} 
\usepackage{natbib}  
\usepackage{caption} 
\usepackage{algorithm}
\usepackage{algorithmic}

\usepackage{newfloat}
\usepackage{listings}
\DeclareCaptionStyle{ruled}{labelfont=normalfont,labelsep=colon,strut=off} 
\floatstyle{ruled}
\newfloat{listing}{tb}{lst}{}
\floatname{listing}{Listing}

\usepackage{subfiles}
\usepackage{csquotes}
\usepackage[dvipsnames]{xcolor}
\usepackage{graphicx}
\usepackage{booktabs}

\begin{document}

\title{Scope of Large Language Models for Mining Emerging Opinions in Online Health Discourse}
\author{
  Joseph Gatto, Madhusudan Basak, Yash Srivastava, Philip Bohlman, Sarah M. Preum\\
  \textmd{Department of Computer Science, Dartmouth College}\\
  \textmd{\{joseph.m.gatto.gr, sarah.masud.preum\} @ dartmouth.edu}
}
\maketitle


\begin{abstract}
    In this paper, we develop an LLM-powered framework for the curation and evaluation of emerging opinion mining in online health communities. We formulate emerging opinion mining as a  pairwise stance detection problem between (title, comment) pairs sourced from Reddit, where post titles contain emerging health-related claims on a topic that is not predefined. The claims are either explicitly or implicitly expressed by the user. We detail (i) a method of \textit{claim identification} --- the task of identifying if a post title contains a claim and (ii) an opinion mining-driven evaluation framework for stance detection using LLMs. 
    
     We facilitate our exploration by releasing a novel test dataset, \underline{L}ong \underline{C}OVID-Stance, or LC-stance, which can be used to evaluate LLMs on the tasks of claim identification and stance detection in online health communities. Long Covid is an emerging post-COVID disorder with uncertain and complex treatment guidelines, thus making it a suitable use case for our task. LC-Stance contains long COVID treatment related discourse sourced from a Reddit community. Our evaluation shows that GPT-4 significantly outperforms prior works on zero-shot stance detection. We then perform thorough LLM model diagnostics, identifying the role of claim type (i.e. implicit vs explicit claims) and comment length as sources of model error. 
\end{abstract}

\section{Introduction}

Characterization of public opinion on complex topics is often a crucial part of many decision making processes. For example, individuals explore aggregated product reviews or aspect-based sentiment before online purchase, or public health organizations assess public opinion to inform health policy design \cite{Ireland-2018-Application}. A critical yet under-explored use-case of public opinion characterization is found in personal health management. People now often turn to social media platforms to get feedback on both diagnosis and treatment of conditions like COVID-19, cancer, sleep disorders, or mental health conditions \cite{chen2021social}. The rising demand for peer opinions on personal healthcare decisions is the product of conflicting or insufficient information on complex, uncertain topics with limited and often inaccessible reliable information.  For example, the emergence of long COVID, a disorder where COVID-19-like symptoms persist long after initial recovery from COVID-19, has significant demand for peer support and feedback as long COVID is very difficult to diagnose with insufficient research on treatment options \cite{doi:10.1080/23744235.2021.1924397}. 

\begin{figure}[!t]
    \centering
    \includegraphics[width=\columnwidth]{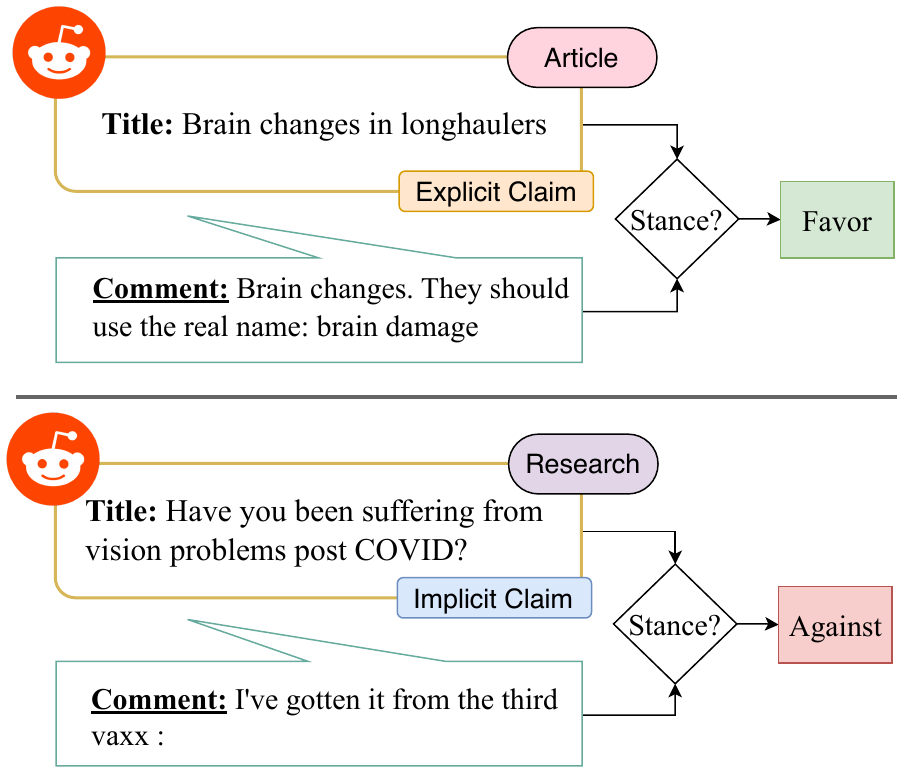}
    \caption{Example post title and comment pairs from r/covidlonghaulers displaying flairs (Article/Research), claim type (Implicit/Explicit), and stance label (In-Favor/Against/None). In the ``Favor" sample (top), the comment supports the claim in the title that long COVID causes brain changes. In the ``Against" sample (bottom), the title and comment disagree on the source of vision problems (i.e. long COVID vs vaccine).}
    \label{fig:example_data}
\end{figure}

In this paper, we aim to characterize emerging public opinion on different long COVID treatments and diagnosis options through analysis of online health discourse. Specifically, we utilize data collected from a relevant Reddit community (i.e., r/covidlonghaulers). Characterization of such online discussion has widespread implications for both personal and public health. (i) At the user level, aggregation of opinion on emerging health topics may help users make more informed decisions and become less reliant on individual posts or comments. (ii) From a public health perspective, this analysis can help researchers study complex, uncertain conditions like long COVID. For example, if many users are in agreement about the efficacy of a self-treatment option, this may provide a useful signal for hypothesizing long COVID treatment research. Conversely, automatic detection of widespread support for dangerous self-treatment options may inform public health education materials.

Our goal is to develop an opinion mining framework that is scaleable and useful in a real-world setting. The modeling of emerging topics in health can prove challenging due to limited data availability. \textbf{Large Language Models (LLMs) have great potential to benefit social opinion mining research} given their strong zero-shot and few-shot reasoning capabilities \cite{gpt3}. Thus in this study, we thoroughly explore the capacity of LLMs to characterize user opinion in online health forums. Specifically we want to find the scope and blindspots of LLMs on this challenging task.

Prior works in this space are either inflexible to emerging topics (i.e. topics not yet represented in a model's training data) \cite{Barbieri-2020-Tweeteval} or use manual data curation, limiting the scalability of such systems \cite{covidlies, Gorrell-2019-Semeval}. We address the issue of flexibility by viewing opinion mining through the lens of \textit{stance detection} --- where a model decides if a health text is In-Favor, Against, or Neutral with respect to a health claim. This framework allows us to quantify public perception of emerging information in a topic-agnostic manner, as the health claim in question is not pre-defined, but is part of the input. We address the issue of scalability by introducing LLMs for automation of the health claim identification process. 

Our formulation is challenging in that it aims to compute the stance between \textit{pairs of online health texts}, where the claim itself can be \textit{deeply implicit}. Such a formulation requires a model to have significant world knowledge, making LLMs well-suited to this task. We thus conduct experiments through the lens of two research questions. \textbf{RQ1:} Can LLMs identify claims in online health texts? and \textbf{RQ2:} How well can LLMs infer the stance of online health texts?

Unfortunately, there is no dataset to evaluate performance of LLMs in this context. Thus, we construct a novel dataset, \underline{L}ong \underline{C}OVID-Stance, or \textbf{LC-stance}.
LC-Stance contains two layers of annotation: (i) \textbf{Claim Identification}: LC-Stance contains 150 Reddit post titles annotated for whether or not they contain a health claim. Claim identification is used to evaluate the quality of our LLM-in-the-loop data curation pipeline. The subtask of claim identification has important implications for large-scale curation of health claim datasets. (ii) \textbf{Stance Detection}: LC-Stance contains 400 human-annotated (title, comment) pairs sourced from Reddit, where each title contains a health claim about emerging long COVID news/research. Claims in LC-Stance are sourced using a novel LLM-in-the-loop pipeline which uses LLMs to scale the collection of claim-driven post titles for human annotation. LC-Stance additionally features unique annotation for fine-grained analysis such as performance on implicit vs. explicit claim types. We benchmark a diverse set of LLMs as well as encoder-only baselines on LC-Stance. We then perform thorough model diagnostics to motivate future work. \\

\noindent We summarize our contributions below: 

\begin{enumerate}
    \item We outline a generalizable LLM-powered data curation and evaluation framework for stance detection in the context of emerging opinion mining. To the best of our knowledge, we are the first to use LLMs to mine emerging opinions in online health texts in an end-to-end manner, demonstrating the capacity of LLMs to both identify implicit and explicit user-generated claims and then perform topic-agnostic pairwise stance detection.

    \item We release \textbf{LC-Stance}, a novel \textbf{test dataset} sourced from long COVID forums on Reddit. LC-Stance contains annotation to evaluate LLMs on both stance detection and claim identification. Following prior works \cite{srivastava2022beyond} we develop a dataset which facilitates evaluation in a zero or few-shot learning setting, where no domain-specific training or fine-tuning is required. All samples additionally contain annotation for claim type (i.e. is the claim implicit or explicit), enabling low-level analysis of LLM error on LC-Stance.

    \item We benchmark 3 popular LLMs, namely, Llama2, GPT-3.5, and GPT-4, on LC-Stance, identifying the best prompting strategies for each task. We perform thorough model diagnostics, identifying the role of claim type (implicit vs explicit claims) and comment length as sources of error. We find that LLMs are significantly more robust to variations in claim type and text length compared to prior works.

\end{enumerate}

\section{Related Work}

\begin{table}[]
\small
\centering
\begin{tabular}{@{}lc|c|c@{}}
\toprule
 & \textbf{Full Data} & \textbf{Implicit} & \textbf{Explicit} \\
\textbf{Unique Titles} & 74 & 29 & 45\\
\textbf{Title-Comment Pairs} & 400 & 206 & 194\\
\midrule

\textbf{Title Word Length} & & & \\
Maximum     & 41      & 20      & 41       \\
Average     & 14.14   & 10.80   & 17.68     \\
\textbf{Comment Word Length} & & & \\
Maximum     & 544     & 518     & 544       \\
Average     & 48.34   & 51.51   & 44.96     \\
 \midrule
 & \textbf{Favor} & \textbf{Against} & \textbf{None}\\
\textbf{Label Distribution} & 188 & 109 & 103\\\bottomrule
\end{tabular}%
\caption{LC-Stance dataset statistics summary.  }
\label{table:dataset}
\end{table}
\subsection{Stance Detection on Social Media} 
Stance detection is the process of identifying a person’s stance or viewpoint towards a particular context or target. This target can be a noun phrase (e.g., “Donald Trump” or “Climate Change”) or a claim. The stance detection problem has gained recent popularity, particularly following the publication of the \cite{mohammad2016semeval} dataset containing stance annotation for Twitter data. Since then, numerous studies have been performed for detecting the stances on social media data,  tackling various tasks such as gauging the opinions of general users \cite{almadan2022userbased,cotfas2021unmasking}, verifying rumors \cite{tian2020early}, and detecting misinformation \cite{hardalov2022survey}. Initial works on stance detection focused on in-target stance detection (where the set of targets in the test data contains the same set of targets as the train data) \cite{mohammad2016semeval,li-etal-2021-p,Gatto-2023-Chain}  or cross-target stance detection (where the target sets in train data and test data are different but from the same domain) \cite{mohammad2016semeval}. More recently, zero-shot stance detection \cite{allaway-mckeown-2020-zero,zhao2023ez,cruickshank2023use}, which involves test data containing the set of targets from domains different than the training data, has garnered attention due to its applicability to data from unexplored domains. 

Our dataset differs from most data in prior works in the following ways: (i) Our dataset uses pairs of online health texts for stance detection. To the best of our knowledge, all other works use either manually curated \cite{covidlies} or human-written \cite{zhao2023ez} data in the stance creation pipeline. (ii) To the best of our knowledge, this is the first stance dataset exclusively modeling online health text pairs sourced from Reddit. (iii) We provide annotation for claim type (implicit vs explicit) for each claim in our dataset. To the best of our knowledge, most prior works do not have annotation for deeply implicit claims. 

One related work in this regard is \cite{glandt-etal-2021-stance}, where they identify if a tweet has implicit stance/opinion towards a pre-defined topic. However, our annotation differs in that our claims are themselves implicit, the opinion of the Reddit user is not what contains explicit/implicit annotation. Another notable work is \cite{Gorrell-2019-Semeval}, where they model rumors using pairs of social media texts.
However, our focus is on modeling stance on emerging claims rather than rumors related to general news. Additionally, our work characterizes claims as explicit vs implicit. 

Opinion mining and sentiment analysis (identifying the user's sentiment towards an entity) \cite{medhat2014sentiment} are also relevant to stance detection. However, we are not focusing on determining the specific sentiment of a user rather their stance with respect to a claim.

\subsection{Fact-checking and Misinformation Detection }
Checking the veracity of a claim is a long-standing problem. Broadly, fact-checking can be divided into three sub-problems - Claim Identification, Evidence Retrieval, and Claim Verification. In the Claim identification, a check-worthy claim is identified from the input text (e.g., a social media post). Then, the relevant and trustworthy evidence set is extracted from the external knowledge bases, known as evidence retrieval. Finally, the veracity of the claim is determined based on the retrieved set of evidence. 

Numerous NLP-based approaches have been employed to solve each sub-problem individually and the full problem as an end-to-end pipeline. A variety of methods, including multi-class classification \cite{Patwari-2017-Tathya}, Distant Supervision \cite{vlachos-2015-identification} and Transformer-based models \cite{Nakov-2022-Overview} have been employed for claim identification task. From simple TF-IDF-based document retrieval \cite{Thorne-2018-Fever} to Graph attention networks \cite{Hu-2023-Unifee} and BERT-based retrieval \cite{Soleimani-2020-Bert} methods were used for evidence retrieval. On the other hand, according to \cite{Das-2023-The-state}, a number of approaches consider claim verification as a binary classification problem \cite{Nakashole-2014-Language,Papat-2018-Declare} while others consider it as a multi-class problem \cite{Augenstein-2019-Multifc,Shu-2020-Fakenewsnet}. Claims can be also be verified using the stances of the corresponding evidences \cite{Deka-2023-Multiple}. Finally, there are some works that tried to solve the fact-checking problem using an end-to-end pipeline \cite{Hassan-2017-ClaimBuster,Li-2018-An-end,Ahmed-2021-Fact}.

Our work differs from such works as the fact-checking literature demands the claim to be explicit while we are focusing on working with both implicit and explicit claims. Also, we are not trying to fact-check the claims; instead, we want to characterize user opinions against any claim, which can come from personal opinions or experiences.

\subsection{LLMs for Online Health Discourse Analysis} 
Due to their complex language understanding capability, LLMs have been used to deal with informal online health texts in diverse health conditions, e.g., mental health \cite{Xu-2023-Mental-LLM,Qin-2023-Read}, COVID-19 \cite{Tekumalla-2023-Leveraging}, etc. LLMs have been used for various purposes, including data annotation \cite{Tekumalla-2023-Leveraging}, text classification \cite{Jiang-2023-Balanced,Xu-2023-Mental-LLM}, explanation generation \cite{Jiang-2023-Balanced}, and chat-based support systems \cite{Qin-2023-Read}. Except for a few cases (e.g., information-seeking events \cite{Sharif-2023-Characterizing}), these models have shown promising results. Motivated by their success, we have utilized their prediction and generation power for our work.

\begin{figure}[!t]
    \centering
    \includegraphics[width=\columnwidth]{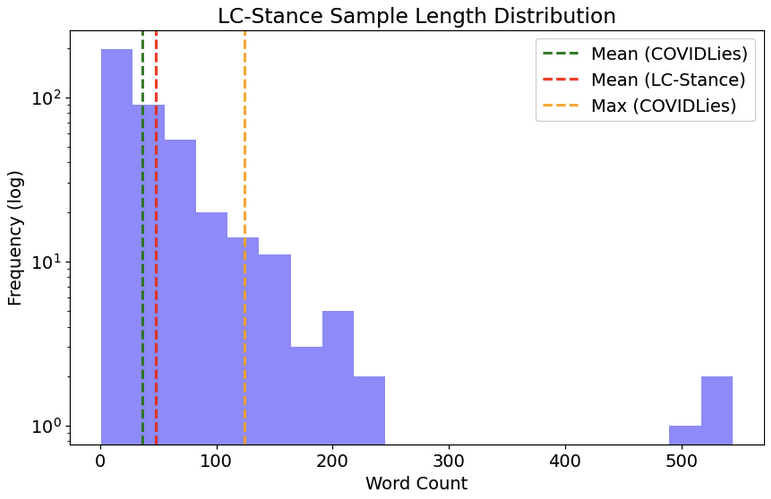}
   \caption{Comparison of the sample length distributions of LC-Stance and COVIDLies, a popular stance dataset sourced from Twitter/X. The mean lengths are similar for both datasets. However, LC-Stance contains significantly longer samples: maximum sample lengths in COVIDLies and LC-Stance are 124 and 544 words, respectively.}
    \label{fig:text_length}
\end{figure}

\section{Problem Formulation}

Consider a social media post $P$ with title $T$, where the discussion thread $D(P)$ has $k$ comments, $D(P) = \{c_1, \dots, c_k\}$. If $T$ contains a \textit{claim} about a healthcare topic, then the \textit{stance} of $c_i$ with respect to $T$ characterizes whether $c_i$ is \textit{in Favor, against, or neutral} about the claim proposed in $T$. Similar formulations have been used in prior works to identify misinformation \cite{covidlies}. This formulation precludes the need to define the problem domain as the opinion of each comment is grounded to the underlying claim of the post title --- thus facilitating topic-agnostic stance detection.

While this formulation can be generalized to any problem domain, we focus on characterizing the emerging opinion of Reddit users on long-COVID as a suitable use case. Consider the top example in Figure \ref{fig:example_data}, where the claim we wish to mine opinions for is ``long-COVID can alter brain structure." In this example, it is difficult to accurately characterize the target and stance of the comment without the context of the post title. In general, this problem is extremely challenging as, unlike prior works, we allow the claim itself to be implicit, requiring a model to have significant world knowledge to properly reason. Our formulation thus allows us to model comments with respect to a claim-driven post title, making opinion characterization topic agnostic. Additionally, our approach promotes semantic analysis of online discussion threads, e.g., how contentious a claim is, without depending on potentially ambiguous and misleading engagement statistics such as up/down votes, number of likes, or comments.


\section{Methods} 

\begin{figure*}[h!]
  \centering
  \includegraphics[width =\linewidth]{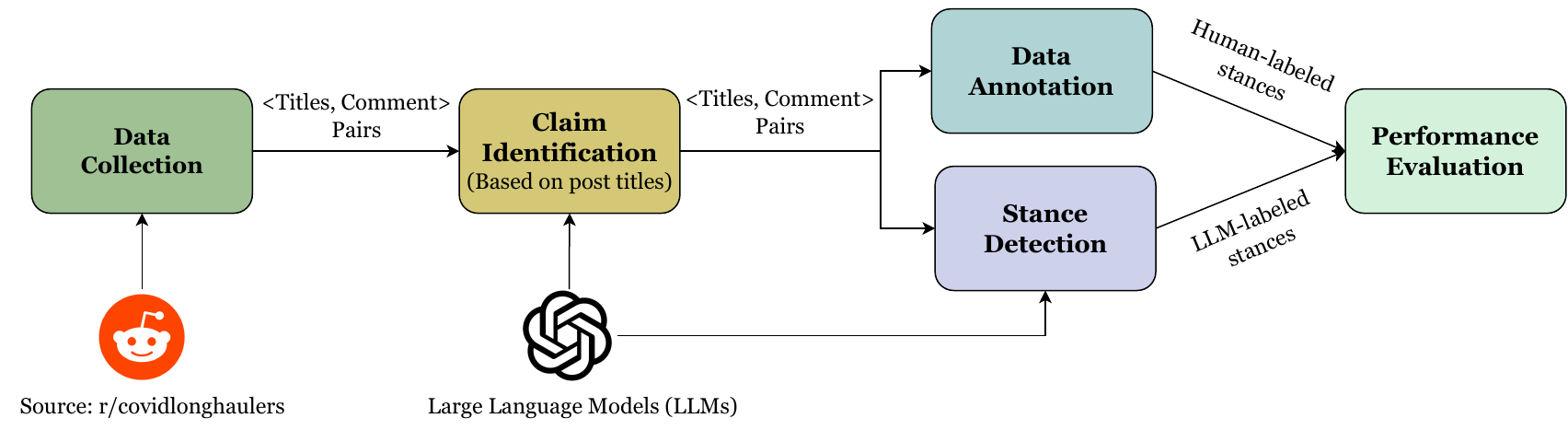}
 \caption{Overall pipeline of our LLM-powered data curation and evaluation framework for topic-agnostic stance detection. First, Reddit data is collected, then titles are filtered using claim identification. Finally, samples are fed to stance classifiers for evaluation.} 
 \label{pipeline}
\end{figure*}

In this section we describe our novel LLM-powered framework for the collection and classification of stance detection data on social media. An overview of our pipeline can be found in Figure \ref{pipeline}.  We first outline our \textit{Data Collection} process, describing source data from r/covidlonghaulers. Then, we detail how we use LLMs to accelerate data curation via \textit{Claim Identification}. Finally, we describe our methods of \textit{Data Annotation} and prompting LLMs for \textit{Stance Detection} of online health texts. 
\subsection{LLM-in-the-Loop Data Curation}

\paragraph{Data Source:} We scrape 2388 posts from the subreddit r/covidlonghaulers between the dates 03/01/2023, 03/31/2023. We choose this subreddit as (i) it is timely (created in July, 2020), (ii) popular (over 50k members as of December 2023) and (iii) focuses on medical topics, including symptoms, and treatment, unlike several other subreddits focusing on political topics related to COVID/Long COVID (LC). Given that we aim to focus our analysis on the opinions of emerging LC topics, we filter the samples to only include posts with either the ``Research" or ``Article" flairs\footnote{A flair is a tag that categorizes the type of post.} as they share emerging LC-related information from peer-reviewed literature, online news articles, blog posts and other online content (e.g., TikTok videos, New York Times, National Institutes of Health, Bloomberg, CNN). After filtering for Research/Article posts, we have 202 unique posts. From this subset, we aim to annotate the stance of comments with respect to post titles which \textbf{express claims} found in emerging long COVID articles and research.

\paragraph{Data Processing for Claim Identification} While posts with Research/Article flairs on Reddit refer to emerging medical research, it is often the case that the claim made in the referenced article is not present in the post title. Even when claims are present, such information can be implicit and may require extensive human effort for annotation. Many prior works have avoided the issue of claim identification by manually-curating claims \cite{covidlies, glandt-etal-2021-stance}. However, we wish to model the stance of \textit{pairs of online health texts} on emerging claims, requiring a more automated approach to post title claim identification. Towards this end, we explore the capacity of LLMs to perform claim identification --- the task of identifying if a post title contains an explicit or implicit health claim. Note that our dataset is unique in that we allow claims to be implicit, as explicit claim mentions are often assumed in prior works \cite{faramarzi2023claim, covidlies, glandt-etal-2021-stance}.

We formulate claim identification as a  binary classification problem, where a post either contains an implicit/explicit claim (Yes), or does not (No). Example explicit and implicit claims can be found in Figure \ref{fig:example_data}. We randomly sample and annotate K=9 post titles not used in LC-Stance for in-context learning of LLMs on claim identification. This leaves 193/202 remaining posts with Article/Research flairs suitable for LLM-powered claim identification. LLMs are well-suited to our task due to the lack of large-scale training data for claim identification in the online health space and their strong capacity to elicit implicit information in texts \cite{Huang-2023-IsChatGPT}. Thus demonstration of LLMs performance on claim identification has significant implications for curating domain-specific claim-driven datasets. 

After claim identification, we are left with 96/193 posts suitable for annotation for stance, i.e., the post titles contain a claim. Next we focus on creating (claim, stance) tuples from these posts. To ensure that comments are in reference to the post content and not other comments, we only consider top-level comments for consideration in LC-Stance. In total, there are 742 first-level comments across these 96 posts. We additionally filter deleted comments, empty comments and comments from Reddit-bots. We also remove posts where the title was incorrectly labeled by LLMs as a claim even when the title did not contain any claim. After filtering, we are left with 679 comments and 74 posts, as some posts got removed due to the above mentioned filtering. From this set, we annotate a random subset of 400 title-comment pairs to produce LC-Stance, where each pair represents a users opinion on a given health claim.

\subsubsection{Dataset Statistics}

A summary of our dataset statistics can be found in Table \ref{table:dataset}. LC-Stance contains 74 unique Reddit post titles, with an average of 5.4 annotated comments per title. Of the 74 titles, 45 contain explicit claims, 29 contain implicit claims. Titles are on average 14 words long, as where comments are on average 48 words long. Unlike Twitter, Reddit comments have no character restrictions, thus LC-Stance is unique in comparison to related works as it requires the modeling longer user-generated texts. For reference, in Figure \ref{fig:text_length} we compare the distribution of post comments in LC-Stance in terms of word length with tweets from COVIDLies, a twitter-based stance dataset. On average, LC-Stance  contains longer samples than COVIDLies and contains a number of significantly longer samples which will help highlight model capacity to process longer online health texts. 

\paragraph{Preparing claim identification test set:} We pair the 74 unique titles in LC-Stance with 76 titles that contain no claim to form a binary classification test set for claim identification. The 76 claimless titles are on average 11 words long, with a max word length of 53.

\paragraph{Data Annotation for Stance Detection:} Each (title, comment) pair was annotated by at least one of the authors, who spent significant efforts to understand relevant background literature on long-COVID, as well as familiarizing themselves with common themes on the r/covidlonghaulers community. Samples were annotated using an iterative annotation process, with frequent reference to authentic information sources and discussion with other annotators.

We note that the choice to use author annotation was necessary for high-quality data annotation in this task as we found \textbf{mTurk-based crowd-sourced annotation to be frequently incorrect}. Specifically, we used a subset of 100 random samples from our 400 sample dataset. Each sample was annotated by three different mTurk workers, and we used majority votes to decide the final label. It results in 33\% of outputs being errors (i.e. wrong stance labels). The poor performance of crowd-sourced annotation is due to their limited knowledge in this topic and lack of incentives to spend additional time to resolve potential confusion and/or lack of knowledge. LC-Stance is thus not suitable for use with crowd sourcing platforms. 

We hope the consideration taken in our annotation process motivates future work to consider if a task is ``crowd-sourceable" as reliance on crowd workers for domain-specific tasks may produce low-quality labels.

While our method ensures that LC-Stance contains high-quality annotation, it is limiting in terms of dataset size, which is subject to future work. However, we note that our dataset size is on-par with related works on LLM evaluation and benchmarking such as Big Bench Hard (BBH) \cite{suzgun2022challenging,srivastava2022beyond}, the subset of the popular Big Bench dataset with highly challenging LLM evaluation sets. The largest evaluation set in BBH, for example, contains only 250 test samples. Thus, LC-Stance contains enough samples for reliable testing of LLMs on this task. 

\subsection{LLM-powered Stance Detection}

Prior works have highlighted the difficulty of obtaining large-scale training data for stance detection from online health communities \cite{covidlies}. Existing solutions to this problem often leverage Natural Language Inference (NLI) data to build zero-shot classifiers \cite{covidlies}, however such data is both out-of-domain and built for sentence-level tasks. We thus thoroughly investigate the scope of zero-shot, few-shot, and chain-of-thought LLM prompting strategies on the LC-Stance dataset. We believe LLMs are well-suited to model LC-Stance for the following reasons. (i) As shown in \cite{chae2023large}, LLMs show strong zero-shot performance on social media-based text classification tasks. (ii) Texts in LC-Stance can be upwards of 544 words long. Thus the capacity of LLMs to model long texts \cite{Wang-2023-Document} makes them well-suited to deal with lengthy posts common to Reddit. (iii) LLMs have shown to substantially outperform text encoders in various domain-specific tasks \cite{gatto2023text}, thus they have strong potential to perform well in online health text modeling. (iv) LLMs have demonstrated strong ability to identify implicit semantics of texts \cite{Huang-2023-IsChatGPT}, thus making them well-suited to model implicit claims in LC-Stance. Please refer to the supplementary materials to review prompts used in our LLM experiments.

\section{LLM Evaluation Setup}
Our evaluation on LLMs for the curation and classification of LC-Stance aim to address the two key research questions (RQs). We evaluate three different LLMs for each RQ: Llama2-7b \cite{touvron2023llama},  GPT-3.5, and GPT-4. We believe this to be a representative set of LLMs as both small open-source models like Llama2 and large closed domain models like GPT-3.5 and GPT-4. Also, GPT-3.5 and GPT-4 are larger than most of the contemporary LLMs (thus likely to show emergent abilities) and widely used in  computational social sciences \cite{Zhu-2023-Ghatgpt}.

\begin{table}[!t]
\centering

\begin{tabular}{@{}lccc@{}}
\toprule
\textbf{Model}      & \textbf{P} & \textbf{R} & \textbf{F} \\ \midrule
\textbf{Out-of-Domain Baseline} & & & \\
ClaimDeBERTa        & 0.63       & 0.63          & 0.63    \\
\midrule
\textbf{Zero-Shot LLM} & & & \\
Llamab2-7b          & 0.75       & 0.52       & 0.37       \\
GPT-3.5-Turbo       & 0.65       & 0.65       & 0.65       \\
GPT-4               & 0.77       & 0.69       & 0.66          \\\bottomrule
\textbf{Few-Shot LLM} & & & \\
Llamab2-7b          & 0.80       & 0.67       & 0.63       \\
GPT-3.5-Turbo       & 0.79       & 0.79       & 0.79       \\
GPT-4               & 0.75       & 0.68       & 0.65          \\\bottomrule
\end{tabular}%
\caption{\underline{P}recision, \underline{R}ecall, and \underline{F}1 scores on claim identification.GPT-3.5-Turbo performs best with a macro F1 score of 0.79. }
\label{table:claim}

\end{table}

\begin{table*}[!ht]
\centering
\small
\begin{tabular}{l|ccc|ccc|ccc|ccc}
\toprule
 & \multicolumn{3}{c}{Macro} & \multicolumn{3}{c}{In-Favor} & \multicolumn{3}{c}{Against} & \multicolumn{3}{c}{Neutral} \\
\cmidrule(lr){2-4} \cmidrule(lr){5-7} \cmidrule(lr){8-10} \cmidrule(lr){11-13}
 & P & R & F1 & P & R & F1 & P & R & F1 & P & R & F1 \\
\midrule
\textbf{Zero-Shot NLI} &  &  &  &  &  &  &  &  &  &  &  &  \\
BART-large             & 0.39 & 0.26 & 0.27 & 0.00 & 0.00 & 0.00 & 0.36 & 0.51 & 0.42 & 0.82 & 0.26 & 0.40 \\
RoBERTa-large             & 0.37 & 0.34 & 0.25 & 0.01 & 0.33 & 0.01 & 0.30 & 0.43 & 0.35 & 0.80 & 0.26 & 0.39 \\
PubMedBERT             & 0.40 & 0.42 & 0.38 & 0.29 & 0.56 & 0.38 & 0.35 & 0.42 & 0.38 & 0.56 & 0.27 & 0.37 \\
PubMedBERT + STS             & 0.43 & 0.43 & 0.41 & 0.35 & 0.62 & 0.45 & 0.38 & 0.34 & 0.35 & 0.57 & 0.35 & 0.43 \\
\midrule
\textbf{Zero-Shot LLM} &  &  &  &  &  &  &  &  &  &  &  &  \\
Llama2-7b              & 0.49 & 0.51 & 0.50 & 0.74 & 0.63 & 0.68 & 0.39 & 0.54 & 0.45 & 0.36 & 0.36 & 0.36 \\
GPT-3.5-Turbo          & 0.49 & 0.58 & 0.44 & 0.43 & 0.81 & 0.56 & 0.20 & 0.61 & 0.30 & \textbf{0.84} & 0.33 & 0.47 \\
GPT-4                  & 0.61 & 0.64 & 0.59 & 0.53 & \textbf{0.82} & 0.64 & 0.52 & 0.70 & 0.60 & 0.77 & 0.40 & 0.52 \\
\midrule
\textbf{Few-Shot LLM} &  &  &  &  &  &  &  &  &  &  &  &  \\
Llama2-7b             & 0.36 & 0.49 & 0.29 & \textbf{0.94} & 0.48 & 0.64 & 0.08 & \textbf{0.75} & 0.15 & 0.06 & 0.25 & 0.09 \\
GPT-3.5-Turbo         & 0.58 & 0.60 & 0.55 & 0.40 & 0.81 & 0.54 & 0.64 & 0.60 & 0.62 & 0.69 & 0.37 & 0.48 \\
GPT-4                 & \textbf{0.71} & \textbf{0.72} & \underline{\textbf{0.71}} & 0.81 & 0.81 & \textbf{0.81} & 0.72 & 0.74 & 0.73 & 0.60 & 0.60 & \textbf{0.60} \\
\midrule
\textbf{Chain-of-Thought LLM} &  &  &  &  &  &  &  &  &  &  &  &  \\
Llama2-7b        & 0.47 & 0.51 & 0.48 & 0.74 & 0.61 & 0.67 & 0.35 & 0.60 & 0.44 & 0.33 & 0.32 & 0.33 \\
GPT-3.5-Turbo    & 0.61 & 0.63 & 0.60 & 0.61 & \textbf{0.82} & 0.70 & 0.51 & 0.64 & 0.57 & 0.71 & 0.42 & 0.53 \\ 
GPT-4            & 0.70 & 0.70 & 0.70 & 0.81 & 0.76 & 0.78 & \textbf{0.74} & 0.74 & \textbf{0.74} & 0.53 & \textbf{0.61} & 0.57 \\
\bottomrule
\end{tabular}
\caption{Performance of various models on LC-Stance for topic-agnostic stance detection. Best performing scores for each metric are bold. The best overall model performance is underlined. We find that GPT-4 with few-shot prompting out-performs all other baselines. }
\label{tab:results}
\end{table*}

\subsection{RQ1: Can LLMs Identify Claims in Online Health Texts?} We evaluate LLMs on the 150 Reddit post titles, annotated for presence of a health claim. Recall that 74 claims in this dataset are from LC-Stance, and the remaining 76 are claimless samples sourced to facilitate this task. We evaluate zero and few-shot prompting of LLMs on this dataset. 

We additionally benchmark the performance of an out-of-domain claim identification model in effort to explore the need for domain-specific solutions. Specifically, we use a publicly available DeBERTa-v2 \cite{he2021deberta} model fine-tuned  on the ClaimBuster dataset 
\cite{arslan2020claimbuster} \footnote{huggingface.co/whispAI/ClaimBuster-DeBERTaV2}, which contains annotation for if political texts (a) contain check-worthy factual claims, (b) unimportant factual claims, or (c) no factual claim. We consider samples classified as either \textit{check-worthy factual claims} or \textit{unimportant factual claims} as claims in our evaluation, with \textit{no factual claim} mapped to no claim. We note that while this label space differs from ours, as claims in LC-Stance can be non-factual or unverifiable, little data similar to our problem exists in the literature. We denote this model as ClaimDeBERTa. We report the precision, recall, and F1 score for each experiment. 

\subsection{RQ2: How well Can LLMs Infer the Stance of Online Health Texts? }  To answer this question, we benchmark each LLM on the LC-Stance dataset using three prompting strategies: zero-shot, few-shot, and Chain-of-Thought (COT) \cite{NEURIPS2022_9d560961} prompting. We demonstrate the need for LLMs on this task by comparing model performance to prior works on zero-shot stance detection. As demonstrated in \cite{covidlies}, Natural Language Inference (NLI) models are an effective zero-shot baseline for stance detection tasks. To evaluate NLI on LC-Stance, we align our stance labels with NLI labels: \textit{In-Favor} to \textit{Entailment}, \textit{Against} to \textit{Contradiction}, and \textit{Neutral} to \textit{Neutral}. 

We evaluate three different pre-trained model types for zero-shot NLI. (i) BART-large \cite{lewis-etal-2020-bart} fine-tuned on MNLI (ii) RoBERTa-large \cite{roberta} fine-tuned on MNLI  (iii) PubMedBERT \cite{Deka-2023-Multiple,pubmedbert} fine-tuned on both MNLI and MedNLI \cite{romanov-2018-lessons}, which contains clinical texts annotated for the NLI task. We additionally run a fourth baseline inspired by \cite{covidlies}, where we combine NLI models with semantic textual similarity (STS) to improve performance. Specifically, we consider (title, comment) pairs with less than 0.4 similarity score to be Neutral, and then allow the NLI model to determine if the pair is more likely to be entailment (\textit{in favor}) or contradiction (\textit{against}). Our experiment denoted PubMedBERT + STS uses this inference strategy.

For each model, we report the precision, recall, and F1 score. We additionally report the class-wise performance of each metric. Overall model performance is determined by the Macro F1 across all classes. Note that for each LLM experiment in the paper, we set the temperature hyperparameter to 0 for reproducibilitiy. For open-source models such as Llama2-7b we utilize tools such as Huggingface Transformers \cite{huggingface} and LangChain\footnote{https://github.com/langchain-ai/langchain}. Experiments run locally (i.e. not using API services) were all ran on the Google Colab computing platform.

\section{Results}

\subsection{RQ1: Can LLMs identify claims in online health texts?}

Our claim identification results can be found in Table \ref{table:claim}. First, we highlight that our out-of-domain baseline model, ClaimDeBERTa, achieves an F1 score of 0.63. This score is competitive with our zero-shot LLM experiments. However, our best performing few-shot prompting method significantly out-performs this baseline by 16 F1 points. Thus, while ClaimDeBERTa has non-trivial performance on this task, there is still great need for domain-specific solutions to this problem. 

LLMs improve performance on claim identification, as they are highly efficient few-shot learners \cite{gpt3}, even for domain-specific health texts \cite{agrawal-etal-2022-large}. Our highest performing model, GPT-3.5, achieves an F1 score of 0.79. For GPT-3.5 and Llama2-7b, we find that few-shot prompting significantly out-performs zero-shot prompting. However, GPT-4 struggles to leverage few shot examples. Due to the opaque nature of GPT-3.5/4, it is difficult to say with certainty why the updated GPT-4 model under-performs it's predecessor. Explanations for this phenomena may include differences in either training data or instruction tuning strategy.

\textbf{Implications}: Our experiments verify LLMs usefulness as a \textit{claim data curation tool} to reduce manual effort in claim identification in online discourse. 
By using our approach to filter through a large number of unannotated samples, we can reduce the human effort to read and reject the annotation of claim-less texts during the annotation process, increasing the efficiency in claim data curation for different relevant tasks, such as, stance detection \cite{faramarzi2023claim}, misinformation detection \cite{Kolluri-2022-Poxverifi}, and fact-checking \cite{Patwari-2017-Tathya}.

\subsection{RQ2: Can LLMs Infer the Stance of Online Health Texts? }
The results of our stance detection experiments can be found in Table \ref{tab:results}. We find that the NLI baselines used in prior works are not sufficient to model LC-stance, with the highest-performing model achieving an F1 score of 0.41. However, we highlight the significant boost in performance when employing PubMedBERT, which is trained on health texts, when compared to RoBERTa-large, which has no domain-specific training. This suggests that LC-Stance requires more domain knowledge than prior work on stance for online health texts, where health text pre-training had little effect on performance \cite{covidlies}. 

Our LLM baselines significantly outperform all NLI baselines, suggesting that LLMs are better-suited for the task of zero/few-shot topic-agnostic stance detection of online health texts. We find that few-shot prompting with GPT-4 yields the best performance, with a macro F1 score of 0.71. When compared to zero-shot prompting, it is evident from our results that in-context demonstrations of stance detection are helpful, as GPT-4, for example, shows a 12 point increase in F1 score with few-shot prompting. The inclusion of COT reasoning sporadically boosts performance (e.g., macro-F1 scores of Llama2 and GPT-3.5).

\textbf{Implications}: LLMs provide significant improvement over prior solutions to zero/few-shot stance detection of online health texts. However, there is significant room for improvement, with the top performing model achieving a 0.71 F1. We explore sources of model in the next section to motivate future innovation on this task.

\section{Model Diagnostics}

LC-Stance contains unique characteristics which may be sources of model error. In this section, we explore two potential drivers of performance degradation from our experiments. First, we identify the relationship between \textit{claim type} and model performance. i.e. How does performance change when the claim is implicit vs explicit? This question applies to both claim identification and stance detection. Additionally, we asses the impact of \textit{comment length} when modeling stance between post title and post comment. 
\begin{figure}
    \centering
    \includegraphics[width=\columnwidth]{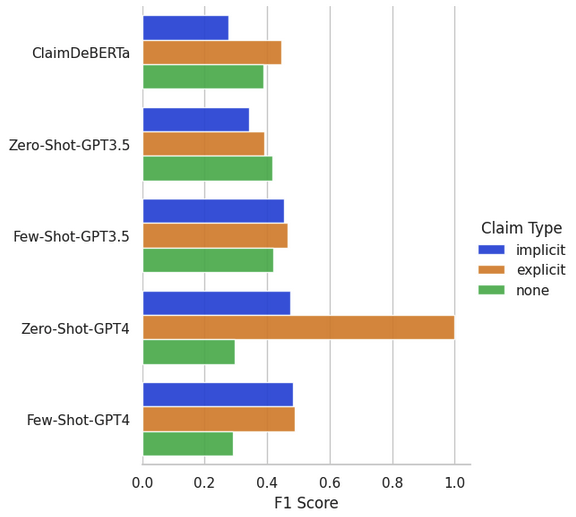}
    \caption{Claim identification F1 scores for each claim type. We find that LLMs are robust to processing varying claim types, with GPT-3.5 exhibiting best overall performance. }
    \label{fig:imp_exp_lcclaim}
\end{figure}

\subsection{Explicit vs. Implicit Claims} 
Recall that explicit claims are substrings of the post title, as where implicit claims are not directly stated but can be understood through context. In this analysis, we re-compute the macro F1 score for each model after binning samples based on their claim type.

\subsubsection{Claim Identification: } 

In Figure \ref{fig:imp_exp_lcclaim} we plot the per-claim performance of ClaimDeBERTa as well as the top-4 performing experiments on claim identification. Notably absent from this list is Llama2-7b, which was the lowest performing model in our experiments. We find that ClaimDeBERTa achieves competitive performance on explicit and none type claims, but struggles to predict implicit claim types. This is because ClaimDeBERTa is, by definition, limited to identifying factual claims. This conflicts with implicit claim types, which can often be non-factual accounts from personal experiences. This result demonstrates that existing large-scale claim identification datasets do not support the modeling of implicit claims. 

GPT-3.5, however, is extremely robust to claim type, as there is little variation in performance across all claim types for both zero and few-shot prompting methods. Thus, for GPT-3.5, claim type is not source of error. 

We find that GPT-4, unlike other models, shows extreme bias towards prediction of the positive class (i.e. the title contains a claim). Thus, performance on predicting the ``none" class is low, while prediction of both implicit and explicit claims is comparatively high. In fact zero-shot GPT-4 correctly identifies every explicit claim in our evaluation set. One explanation for this is that explicit claims in our dataset often contain far more technical vocabulary than implicit claims. Thus, improvements made in GPT-4 vs GPT-3.5 may pertain to greater capacity to model domain-specific texts.

In summary, GPT-3.5 exhibits the best overall performance, with general robustness to claim type. GPT-4, on the other hand, is very sensitive to claim type with strong bias towards implicit claims.

\subsubsection{Stance Detection: } 
\begin{figure}[!t]
    \centering
    \includegraphics[width=\columnwidth]{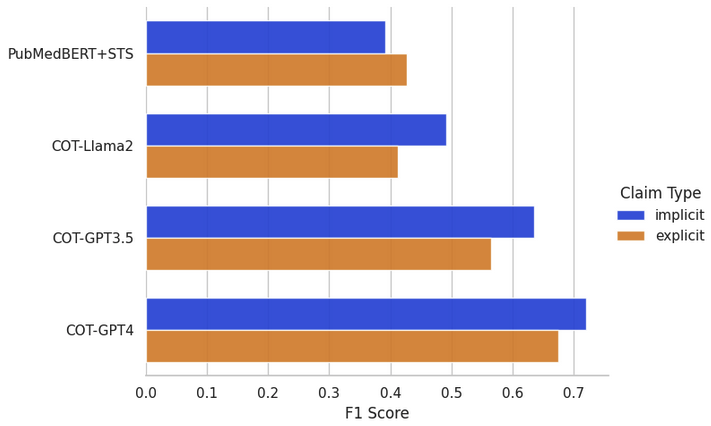}
    \caption{Stance detection performance broken across claim types. We find that all models are generally robust to claim type, with 3/4 exhibiting higher performance on implicit examples. }
    \label{fig:claim_filter_class}
\end{figure}

Figure \ref{fig:claim_filter_class} highlights how performance varies for different claim types. Specifically, we plot the best encoder-only baseline (PubMedBERT+STS) vs each LLM using COT reasoning, which was shown to be the best prompting strategy on average. We find that \textbf{all models show robustness to claim type}, as there is minimal fluctuation between implicit and explicit F1 scores in all experiments. Stance in comments corresponding to titles with explicit claims are more difficult to predict than implicit claims for all LLMs. We hypothesize this occurs due to the larger number of technical medical terms found in explicit claims vs implicit claims. This hypothesis is supported by PubMedBERT+STS being the only model where performance on explicit claims outperforms implicit performance. This is intuitive as PubMedBERT has significant exposure to technical medical terminology during training. Future works may thus wish to perform domain adaptation to enhance LLM performance on LC-Stance. However, overall we find that claim type is not a main driver of error for stance detection.

\subsection{The Effect of Comment Length} 
Long texts have high semantically complexity which may pose challenges to NLP models. However, recent advances in LLMs have shown great improvement on document-level NLP tasks \cite{Wang-2023-Document}. Thus in this analysis, we analyze the sensitivity of stance detection performance given \textit{varying comment lengths}. For each model, we subset samples in LC-Stance by word count \footnote{Word count computed using NLTK word tokenizer. https://www.nltk.org/api/nltk.tokenize.html} and re-compute the macro F1 score. We consider performance at three difference scales: (i) Short comments (between 1 and 50 words) (ii) medium comments (between 50 and 100 words) (iii) long comments (100+ words). This analysis is vital to future works on online health text modeling, as changing policies on popular platforms like Twitter/X have recently removed character restrictions. Thus, it is important that we quantify how robust LLMs are to variations in comment length when performing pairwise stance detection of online texts.

\begin{figure}[t]
    \centering
    \includegraphics[width=\columnwidth]{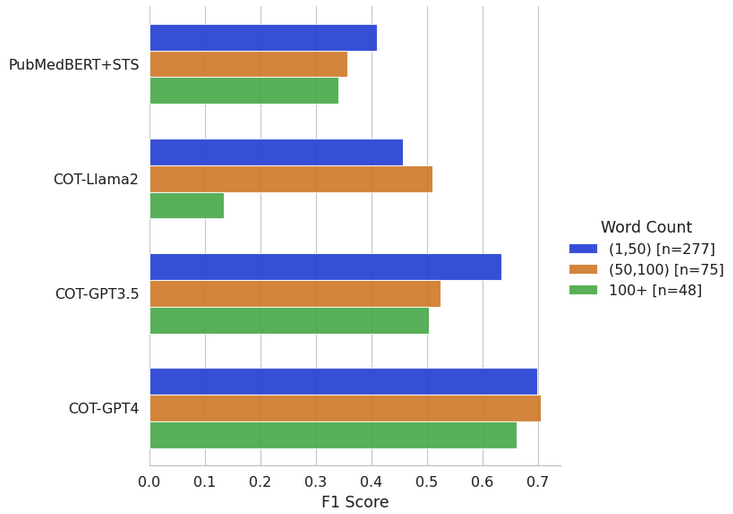}
    \caption{F1 scores on LC-Stance based on comment length. We find that the highest-performing model, GPT-4, exhibits robustness to text length. }
    \label{fig:comment_len}
\end{figure}

Figure \ref{fig:comment_len} shows how LC-Stance performance varies with respect to comment length. We plot the performance of the best encoder-only baseline, PubMedBERT+STS, as well as each LLM using the best average prompting strategy, COT reasoning. We find that all GPT-based LLMs significantly out-perform the best 0-Shot NLI baseline by a significant margin on long texts. Llama2, an LLM with far fewer parameters than GPT-3.5 and GPT-4, shows poor performance on long texts, suggesting modeling of such information may be an ability that emerges at scale. In summary, \textbf{our results show that GPT-3.5 and GPT-4 significantly outperform prior works in this space when processing lengthy texts.}

We notice that both PubMedBERT+STS and COT-GPT-3.5 show higher variability between short and medium texts vs medium and long texts. This suggests that, while it is easier for such models to process short texts, they are in-general robust to increases in text length. Our best performing model, GPT-4, shows very little performance fluctuation in the presence of all word lengths. For example, there is only a 3 point difference in F1 between short and long texts for GPT-4. Our results are promising for future research on LLMs for online health texts, as we show high performance on stance detection which is robust to  post length --- significantly outperforming prior works on zero-shot stance detection of online health texts.

\section{Limitations and Future Work}
One limitation of our work is the sample size of our evaluation sets. While  other works \cite{suzgun2022challenging} have released LLM evaluation sets of comparable size, we wish to evaluate LLMs on larger datasets in the future --- expanding both number of samples and domains used in evaluation. Our data collection pipeline is well-suited towards data collection from other subreddits with flairs similar to the Research/Article flairs used to source LC-Stance. 

We note that the GPT models achieving the best performance on LC-Stance are not free and little is known about the data used to train these models. We additionally note that outside the scope of this work was prompt optimization. Future works may wish to explore intelligent methods of example selection for in-context learning beyond random sampling. Our most successful models had a large number of parameters, likely due to the emergent abilities of LLMs at scale. We will explore how to make smaller LLMs viable for this problem in the future. Additionally, future efforts will focus on extraction of claims from the post body content, which is a far more complex inference task.

\section{Acknowledgements}
Generative AI was used to aid in the construction of tables and plots displayed in this manuscript. All AI-generated materials were carefully reviewed and edited as required by the authors before being included in the paper.

\bibliography{ref}

\newcommand{\answerYes}[1]{\textcolor{blue}{#1}} 
\newcommand{\answerNo}[1]{\textcolor{teal}{#1}} 
\newcommand{\answerNA}[1]{\textcolor{gray}{#1}} 
\newcommand{\answerTODO}[1]{\textcolor{red}{#1}}

\section{Ethics Statement}
All research was conducted under approval from the Institutional Review Board (IRB) at the submitting author's institution. We collect all data from the subreddit r/covidlonghaulers. Reddit was chosen as the data source as it contains anonymous, publicly available user-generated health texts. We ensure that no sample included in LC-Stance contains personally identifying information. To prevent misuse of any user data used in this study, we will require users to sign a data use agreement before accessing LC-Stance. 

A potential negative impact of our dataset is that surfacing divergent opinions on long COVID treatments may cause anxiety among affected individuals. However, the potential benefits of surfacing such information (e.g., creating awareness, encouraging people to contact a professional healthcare provider, inform public health policy research, etc.) outweigh the potential negative impacts.

We note that there may be associated cost in misclassification of claims, such as missing relevant claims. However we anticipate that analyzing a larger number of posts can mitigate this effect as similar claims are often mentioned in multiple posts. Additionally, misclassifying stance may lead to less accurate measure of public opinion.

\end{document}